\documentclass[graybox]{svmult}
\usepackage{algorithm}
\usepackage{algpseudocode}

\usepackage{type1cm}        
\usepackage{makeidx}         
\usepackage{graphicx}        
\usepackage{multicol}        
\usepackage[bottom]{footmisc}
\usepackage{multirow}

\newcommand{\Prob}{\mathrm{Pr}}
\newcommand{\Betadist}{\mathrm{Beta}}

\usepackage{xcolor}  

\usepackage{newtxtext}       %
\usepackage[varvw]{newtxmath}       

\makeindex             

\begin{document}

\title*{Lightweight Change Detection for Adaptive Collective Perception in Dynamic Environments}
\titlerunning{Adaptive Collective Perception in Dynamic Environments}
\author{
  Thiemen Siemensma \and 
  Tiago Travassos \and
  Jiacheng Zhang \and
  Bahar Haghighat
}
\authorrunning{Siemensma et al.}

\institute{
  Faculty of Science and Engineering, University of Groningen, The Netherlands \\
  \email{t.j.j.siemensma@rug.nl, t.j.martins.da.silva.de.oliveira.travassos@student.rug.nl,\\ jiacheng.zhang@rug.nl, bahar.haghighat@rug.nl}
}
%
%
\maketitle

\abstract{Swarm robotics enables scalable inspection through collective perception, where robots infer environmental properties from distributed observations. A common benchmark considers robots exploring black-white tiled environments and determining the majority tile type. While most work studies static environments, dynamic environments remain underexplored.
We study collective perception in dynamic environments where the tile proportion and the spatial correlation of surface patterns may change over time. To enable implementation on resource-constrained robots, we integrate a lightweight change detection mechanism into a Bayesian collective decision-making framework that resets robot beliefs when distribution shifts are detected. The proposed detector requires constant memory and constant per-sample computation, unlike two recent competitive approaches that store the full sample stream, making it suitable for memory-constrained robots. 
We further introduce a method that detects spatially correlated sampling and adapts sampling intervals online to mitigate correlation effects. Our simulation results in Webots show competitive change detection delays, more consistent precision across communication settings, and improved decision accuracy under changing spatial correlations.
}

\section{Introduction}
Robotic inspection has gained significant interest over recent decades \cite{halder2023robots}.
A rising trend is the application of robot swarms to distributed inspection tasks \cite{halder2023robots,Dorigo2021SwarmFuture,ramshanker2026tumblenauts,haghighat2022approach}. 
Robot swarms enable scalable sensing for applications ranging from infrastructure inspection to disaster response \cite{Siemensma2025DistributedRobots, Carrillo-Zapata2020MutualInspection}. 
During such operations, a crucial challenge is collective decision-making based on distributed observations made by individuals in the swarm. 
In infrastructure inspection tasks, this may correspond to determining whether a structure is safe to operate or requires maintenance \cite{Siemensma2026OptimizationRobots}. 

Numerous studies formalize collective decision-making tasks in best-of-$n$ problems, commonly framed as site-selection tasks (analogous to ant nest choice) or as a tile classification task in environments with distinct tile colors \cite{Valentini2017ThePerspectives, Hamann2026TenChallenges}. 
The swarm is tasked to select the best option (e.g. site/tile type) among $n$ options available. 
In particular, the collective perception case for $n=2$ is well studied and is generally formulated in black–white tile environments, where a swarm must identify the majority tile type \cite{Morlino2010CollectiveRobots, Valentini2016CollectiveSwarm}. 
Prominent solutions include consensus-based strategies \cite{Valentini2016CollectiveSwarm, Valentini2015EfficientTrade-Off, Valentini2016CollectiveProblems, Valentini2014Self-organizedModel} and Bayesian frameworks \cite{Ebert2020BayesSwarms, Shan2021DiscreteSharing, Shan2020CollectiveTesting}. 
Across these approaches, a fundamental speed–accuracy trade-off arises: faster decisions typically reduce accuracy, and vice versa \cite{Reina2024Speed-accuracyModeling, Valentini2016CollectiveProblems,Zakir2024MiscommunicationDecision-making, Valentini2015EfficientTrade-Off}. 
The implications of this trade-off are well studied in static environments, however, few studies consider dynamic environments, where surface patterns may change over time \cite{Talamali2021WhenCommunication, Hamann2026TenChallenges, Khaluf2017TheBistability}.
Moreover, the few existing studies primarily focus on changes in the fill ratio, while changes in the surface pattern correlations remain unexplored. Correlation in the surface patterns is known to considerably influence decision-making difficulty \cite{Bartashevich2021Multi-featuredCorrelations, Bartashevich2019BenchmarkingDecision-making}.

In dynamic environments, a key limitation of existing decision-making algorithms is decision lock-in, representing an absorbing decision state that cannot be exited regardless of new observations, thereby preventing decision reversal \cite{Hamann2026TenChallenges, Zakir2022RobotCross-Inhibition, Pfister2022CollectiveEnvironments}. Prior work mitigates lock-in using stubborn agents \cite{Prasetyo2019CollectiveEnvironments}, discarding outdated information \cite{Soorati2020PlasticityLock-ins}, and limiting communication \cite{Talamali2021WhenCommunication,Aust2022TheEnvironments}. Lower sampling rates can further improve reactivity by reducing sample correlation \cite{Aust2022TheEnvironments}. In static environments, sampling rate should correspond to feature size \cite{Ebert2020BayesSwarms, Bartashevich2019BenchmarkingDecision-making}, yet robots have not been shown to estimate and adapt to changing spatial correlations.

Several approaches propose to mitigate decision lock-in by extending the Bayesian decision-making framework Bayes Bot \cite{Ebert2020BayesSwarms}. In Bayes Bots, robots model the fill ratio using a Beta distribution and use the corresponding confidence, called belief, to make final decisions. Within the framework, a hysteresis mechanism that allows flexible decisions by requiring robots to exceed an evidence threshold before deciding was introduced in \cite{Chiu2024OptimizationOptimization}, however, its performance in dynamic environments remains unstudied. The Dynamic Bayes Bots (DBB) algorithm is specifically designed for dynamic fill ratio decision-making by restraining the belief at $p_{\text{lim}}$ and resetting decisions when the belief drops under the threshold $p_{\text{res}}$ \cite{Pfister2022CollectiveEnvironments}. Moreover, robots enhance adaptivity through three types of feedback: (i) a model reset by a more experienced neighbor with an opposite decision, (ii) a model overtake by a more experienced neighbor while indecisive, and (iii) posterior updates from neighbors that share the same final decision. While effective, DBB’s accuracy is constrained by $p_{\text{lim}}$, and resets are slow when the Beta posterior becomes inertia-heavy \cite{Pfister2022CollectiveEnvironments}.

To address these limitations, Dynamic Bayes Bot Change Point Detection (DBBCPD) was proposed \cite{Pfister2023CollectiveEnvironments}, incorporating the change detection algorithm Pruned Exact Linear Time (PELT) \cite{Killick2012OptimalCost}. Robots reset either via PELT detections or through received feedback within a given time period when: (i) encountering $n_1$ resetting neighbors, (ii) if undecided, adopt the decision that is held by a $n_2$ majority of neighbors, or (iii) a decision inversion when $n_3$ neighbors hold an opposite decision. 
DBBCPD improves adaptivity over DBB but must store the full sample stream, requiring $\mathcal{O}(n)$ memory (with $\mathcal{O}(n^2)$ cumulative compute from re-running detection), which renders it infeasible on the memory-constrained (32 kB) Kilobots used in their agent-based simulation study. 
Moreover, the effects of collision avoidance, correlated sampling, and long-range communication (3 body lengths) remain unclear.

As an alternative, computational expenses of change detection can be reduced using control charts, such as CUmulative SUM charts (CUSUM) \cite{Mishra2015ASensors, Crowder2017AnCUSUM}, which are lightweight statistical tools that integrate deviations from an online-estimated in-control mean with $\mathcal{O}(1)$ computational complexity \cite{Lowry1995ACharts, Knoth2004ControlReview}. However, such approaches have not yet been applied to collective perception.

This work improves the adaptivity and computational efficiency of Bayesian dynamic collective perception through three contributions: (i) an $\mathcal{O}(1)$-memory CUSUM fill ratio change detector; (ii) an online estimator of spatial sampling correlation; and (iii) an adaptive sampling strategy that mitigates correlation by adjusting sampling intervals. We evaluate the methods against DBB and DBBCPD in our physics-based Webots framework \cite{Siemensma2026OptimizationRobots}.

\begin{figure}[t!]
    \centering
    \begin{minipage}{0.275\textwidth}
        \centering
        \includegraphics[width=\linewidth]{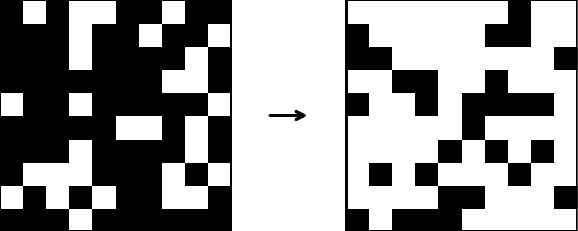}
        (a) Change in fill ratio
    \end{minipage}
    \hfill
    \begin{minipage}{0.275\textwidth}
        \centering
        \includegraphics[width=\linewidth]{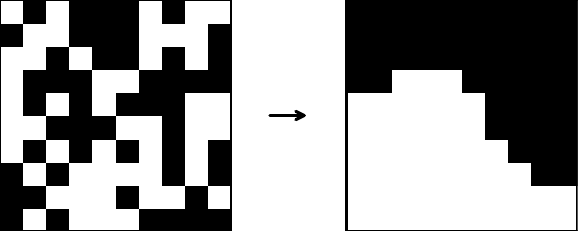}
        (b) Change in correlation
    \end{minipage}
    \hfill
    \begin{minipage}{0.275\textwidth}
        \centering
        \includegraphics[width=\linewidth]{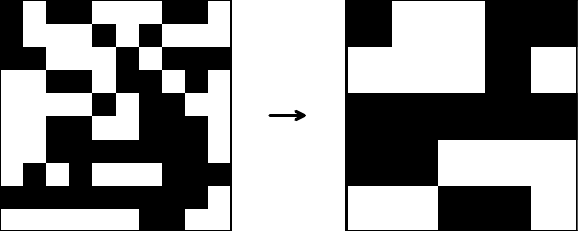}
        (c) Change in feature size
    \end{minipage}
    \caption{We consider three possible changes in our study of dynamic environments:
    (a) fill ratio changes from predominantly black to predominantly white ($f_0=0.3 \rightarrow f_1=0.7$).
    (b) Spatial correlation changes due to clustered tiles.
    (c) Spatial correlation changes due to feature size changes.}
    \label{fig:env_changes}
\end{figure}

\section{Problem Statement}
\label{sec:problem_statement}
We consider a swarm of $N_r$ robots operating in a $1\text{m} \times 1\text{m}$ dynamic environment composed of black and white tiles of size $\ell \text{ m} \times \ell \text{ m}$, where the fill ratio $f$ represents the proportion of white tiles.
As illustrated in Figure~\ref{fig:env_changes}, we consider three environmental changes: (i) changes in fill ratio ($f_0 \rightarrow f_1$), (ii) changes in spatial correlation due to clustering, and (iii) changes in spatial correlation due to feature size variations ($\ell_0 \rightarrow \ell_1$), where $(\cdot)_0$ and $(\cdot)_1$ denote pre- and post-change values, respectively.
The robots are tasked with determining the majority tile type over time while actively estimating and adapting to spatial correlations to enhance decision-making accuracy.

\begin{algorithm}[t]
\caption{Sampling and Decision-Making Strategy}
\label{alg:bayesbot}
\begin{algorithmic}[1]
    \Require $\theta_c,m, \mu_R, \lambda_R, d, d_l = 0, d_f = -1$, $\Betadist(\alpha,\beta) \gets \Betadist(\alpha_0=1, \beta_0=1)$
    \While{ $t<T_{\text{end}}$}
        \State Perform random walk $R(t) \sim \text{Cauchy}(\mu_R, \lambda_R)$
        \If{$R(t) - d_l >d$} \Comment{Sample every $d$ meters}
            \State $C \leftarrow $ Local measurement 
            \State $d_l \gets R(t)$
        \EndIf
        \If{$C$ available } \Comment{Local or received sample available}
            \State $\text{Beta}(\alpha,\beta) \leftarrow \text{Beta}(\alpha + C,\; \beta + (1 - C))$
            \State $p = \Prob\!\left(\text{Beta}(\alpha,\beta) < 0.5\right)$ \Comment{Robot posterior}
            \State $\hat f_0 = \alpha / (\alpha + \beta)$ \Comment{fill ratio estimate}
            \If{\textit{CusumDetector($C, \hat f_0,\theta_c$)}} \Comment{\textbf{Extension 1}: CUSUM detector}
            \State $\text{Beta}(\alpha,\beta) \leftarrow \text{Beta}(\alpha_0, \beta_0)$, $d_f \gets -1$ \Comment{Reset after change detection}
            \EndIf
            \State $d \gets$\textit{CorrelationDetector($C, \hat f_0,m$)} \Comment{\textbf{Extension 2}: Correlation detector}
        \EndIf
        \If{$p > p_c$} \Comment{Majority black tiles}
        \State $d_f = 0$
        \ElsIf{$p < (1-p_c) $} \Comment{Majority white tiles}
        \State $d_f = 1$
        \EndIf
    \EndWhile
\end{algorithmic}
\end{algorithm}

\section{Proposed Algorithm}
\label{sec:algorithm}
We propose two stand-alone extensions to the Bayes Bot framework \cite{Ebert2020BayesSwarms} described in Algorithm \ref{alg:bayesbot}:  
(i) a lightweight fill ratio change detection approach based on CUSUM \cite{Crowder2017AnCUSUM}, and  
(ii) a mechanism to estimate and adapt to spatial correlations.
In our algorithm, we do not exploit the positive feedback mechanism introduced in~\cite{Ebert2020BayesSwarms}, as it leads to decision lock-in through reinforced posterior updates. 

Each robot performs a L\'evy random walk with segments $R \sim Cauchy(\mu_R,\lambda_R)$. 
After every completed segment, or upon detecting a collision within 1.5 body lengths, the robot turns a random angle $\phi \in [-\pi,\pi]$.
During the random walk, the robot uses wheel encoder odometry \cite{Siemensma2025DistributedRobots} to stop every $d$ meters to acquire a binary measurement $C \in \{0,1\}$, where $C=1$ denotes white and $C=0$ black. To reliably infer $C$, each sample is acquired during a $1\,\mathrm{s}$ sampling duration as found in our previous work \cite{Siemensma2026OptimizationRobots}.
Given a pre-change fill ratio $f_0$, each uncorrelated sample follows a Bernoulli distribution with probability $f_0$.
Robots use their own samples and evidence received from neighbors (within communication range) to update their posterior as $\mathrm{Beta}(\alpha + C,\, \beta + (1 - C))$, where $\alpha$ and $\beta$ denote the counts of white and black samples, respectively. By evaluating its belief, i.e. $p = \Pr(\text{Beta}(\alpha, \beta) < 1/2)$, a robot's final decision $d_f$ is set to $1$ if $p < 1 - p_c$, $0$ if $p > p_c$, and $-1$ otherwise.
This decision is kept until a contradicting decision is deemed warranted.

\textit{Detecting fill ratio changes}: 
Under the no-change hypothesis $H_0$ ($C \sim \text{Bernoulli}(f_0)$), the sample probability is estimated as the posterior mean $\hat{f}_0 = \alpha / (\alpha + \beta)$. By cumulatively accumulating sample deviations from this mean, we construct the following CUSUM control chart to detect changes in sample probability (i.e. fill ratio) as:
\begin{subequations}
\label{eq:cusum_chart}
\begin{align}
    S_H^{\,n+1} &= \max \bigl(0, S_H^n + (C - \hat {f}_0 - \delta) \bigr), \label{eq:SH} \\
    S_L^{\,n+1} &= \max \bigl(0, S_L^n - (C - \hat {f}_0 + \delta) \bigr), \label{eq:SL}
\end{align}
\end{subequations}
where $S_H^n$ and $S_L^n$ denote the upper and lower statistic for the $n$-th sample, respectively, and $\delta$ is a parameter controlling sensitivity and including negative drift under $H_0$ \cite{Mishra2015ASensors}. 
The expected increment of the CUSUM chart (i.e., $\Delta S = \pm (C - \hat f_0 \mp \delta)$) equals $\mathbb{E}[\Delta S \mid H_0] = -\delta$ under the no-change hypothesis, and $\mathbb{E}[\Delta S \mid H_1] = f_1 - f_0 - \delta = \Delta f - \delta$ under a change. 
It is common practice to balance upward and downward drift by setting $\delta = \Delta f / 2$, where $\Delta f$ corresponds to the desired detectable change \cite{Mishra2015ASensors}.
A change is detected at time step \(n+1\) according to the decision function \(D(n)\) as:
\begin{equation}
D(n) =
\begin{cases}
1, & \text{if } S_H^{n +1} > \theta_s \;\text{or}\; S_L^{n+1}> \theta_s, \\
0, & \text{otherwise},
\end{cases}
\end{equation}
where $\theta_s$ denotes the CUSUM threshold. When a change is detected, the robot resets its fill ratio model as $\text{Beta}(\alpha, \beta) \leftarrow \text{Beta}(\alpha_0, \beta_0)$, and $d_f \leftarrow -1$.

The CUSUM threshold $\theta_s$ is chosen to achieve a desired false-alarm Average Run Length $\text{ARL}_0$ \cite{Mishra2015ASensors, Crowder2017AnCUSUM}, i.e., the expected number of samples between false alarms caused by random fluctuations in the sample stream.
Since Equation \eqref{eq:cusum_chart} behaves like a one-dimensional random walk with negative drift, the first-passage time till $\theta_s$ satisfies 
$\text{ARL}_0 \sim c \cdot e^{h \cdot \theta_s}$ \cite{Kugler2013UpperDrift}, 
where $c$ and $h$ can be found using Monte Carlo simulations \cite{Mishra2015ASensors}.
Moreover, for a given threshold $\theta_s$, the run length values (RL) converge to the exponential distribution $\text{RL} \sim \exp(\lambda)$, with rate $\lambda = 1 / \text{ARL}_0$ \cite{Tartakovsky2005AsymptoticConstraint}.
Through these relations, we set the $\text{ARL}_0$ for a desired probability $\theta_p$ of reaching $\theta_s$ under a no-change hypothesis within $N_s$ samples using the inverse exponential cumulative distribution function as:
\begin{equation}
    \label{eq:arl0_selection}
    \text{ARL}_0 = \frac{N_s}{- \log (1 - \theta_p)}.
\end{equation}
In this work, the robots compute the sample count $N_s$ online as the product of the moving average sample rate and the total inspection time. 
Finally, since $N_s$ varies with the sample rate, due to variations in communication range and step size, $\theta_s$ is correspondingly adjusted online through the inverse first-passage approximation as:
\begin{equation}
    \label{eq:h_selection}
    \theta_s= \dfrac{1}{h} \log(\text{ARL}_0) - \dfrac{1}{h} \log(c).
\end{equation}
This online approach ensures a false-alarm probability independent of the sample rate, though lower sample rates increase the delay in detecting true changes.

\textit{Detecting correlations}: 
Our second extension targets the detection of spatial correlations, known to reduce decision-making accuracy~\cite{Ebert2020BayesSwarms, Siemensma2026OptimizationRobots, Bartashevich2019BenchmarkingDecision-making, Bartashevich2021Multi-featuredCorrelations}. We mitigate the adverse impacts of these correlations by (i) detecting their presence and (ii) adapting the robot step size $d$ accordingly.
Our approach is based on the robot's posterior mean $\hat f_0$ to compute the expected probability of a one-step flip, i.e., transitioning from black to white or white to black, denoted $\hat p_f$, following:
\begin{equation}
\begin{aligned}
    \hat p_f 
             &= \hat f_0 (1-\hat f_0) + (1-\hat f_0)\hat f_0 = 2 \hat f_0 (1-\hat f_0).
\end{aligned}
\end{equation}
$\hat p_f$ is maximal at $\hat f_0 = 0.5$ and zero when $\hat f_0 \to 0$ or $1$. 
The probability $\hat q_g$ of observing at least one flip in $N_t$ uncorrelated transitions increases with $N_t$ as:
\begin{equation}
    \label{eq:global_flip_prob}
    \hat q_g = 1 - (1 - \hat p_f)^{\,N_t}.
\end{equation}
Similarly, a robot's empirical local flip probability during its last $m$ samples is computed from that personal sample set, denoted $\mathbf{C} \in \{0,1\}^m$. For back-to-back 
samples within this set, the observed one-step flip probability corresponds to
\begin{equation}
 \label{eq:local_flip_prob}
\hat q_l = \frac{1}{m-1} \sum_{i=2}^{m} \mathbf{1}_{\{C^{i} \neq C^{i-1}\}},
\end{equation}
where $\mathbf{1}_{(\cdot)}$ equals 0 if samples are equal, and one otherwise. 
We set $\hat q_l = \hat q_g$ to estimate the number of transitions in a single step $d$ as: 
\begin{equation}
    \label{eq:N_estimate}
   N_t =
    \log\bigl(1 - \hat q_l\bigr) \  / \ \log\bigl(1 - \hat p_f\bigr) \,.
\end{equation}
In the absence of correlation $N_t \rightarrow 1$, whereas $ N_t < 1$ corresponds to locally correlated sampling, where on average, the step $d<\ell$.
Using the current step size $d$ and the value of $ N_t$, the estimated tile size, denoted $\hat \ell $, is therefore given as:
\begin{equation}
    \label{eq:hat_ell}
    \hat \ell =  
    \begin{cases}
        d / (\kappa \  N_t), & \text{if } d \leq \ell,\\
        d, & \text{otherwise,}
    \end{cases}
\end{equation}
where $\kappa = \tfrac{1}{2}\bigl(\sqrt{2}+\ln(1+\sqrt{2})\bigr) \approx 1.148$ is a geometric constant: the mean distance from a tile center to a uniformly random point in the surrounding $3\times3$ tile neighborhood. 
The tile size cannot be estimated when $d > \ell$, since in this case, independent sampling saturates the estimator ($  N_t\rightarrow 1$) and $\hat \ell \rightarrow d$.

\textit{Adapting to correlations}:
We propose updating the step size $d$ every $m$ samples (i.e. every time $\mathbf{C}$ is full) to mitigate the adverse impact of spatial correlations. 
Increasing $m$ improves update accuracy but increases the time between updates.
Since robots can only detect steps that are too small, they adjust $d$ as: (i) increase $d$ when they traverse $\Delta N$ fewer tiles than desired to improve accuracy, and (ii) decrease $d$ when observed correlation is low to increase decision speed. The overall update rule equals:
\begin{equation}
d \gets
\begin{cases}
\min (d /   N_t, 2 \cdot  d), & \text{if }  N_t < 1 - \Delta N  \\
d \left(1 - \eta \cdot \Delta N \right), & \text{otherwise},
\end{cases}
\end{equation}
where $\Delta N$ controls the update rule's sensitivity, and $\eta$ determines the magnitude of the negative drift. We limit the increase of $d$ up to $2\cdot d$ to avoid large increments. Since the decay branch also applies at the target ($N_t\approx1$), $d$ does not converge to a fixed point but instead settles into a bounded oscillation around the tile size.

\begin{figure*}[!t]
    \centering
    \begin{minipage}[t]{0.175\textwidth} 
        \centering
        \includegraphics[width=\linewidth]{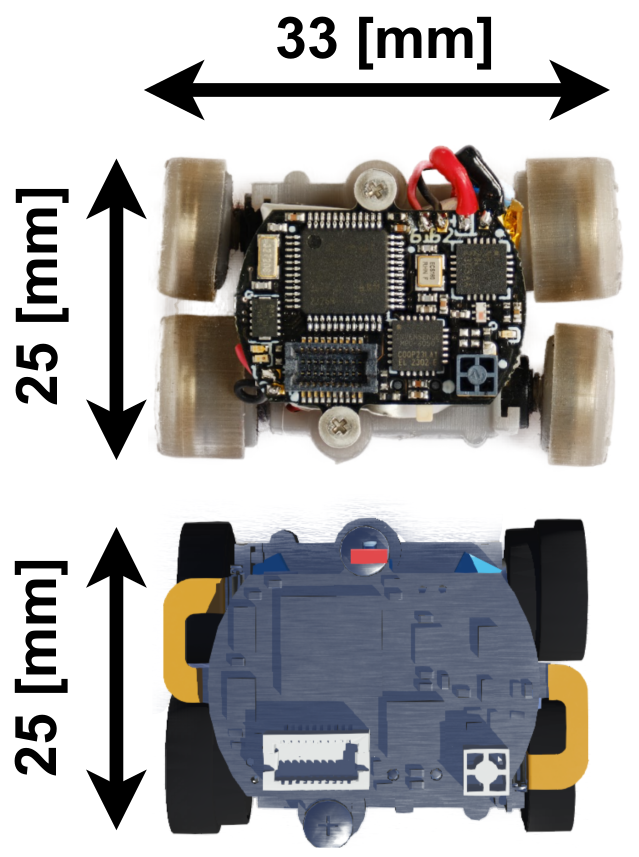}
        \parbox{\linewidth}{\centering (a) Robot model} 
    \end{minipage}
    \hfill 
    \begin{minipage}[t]{0.3\textwidth} 
        \centering
        \includegraphics[width=\linewidth]{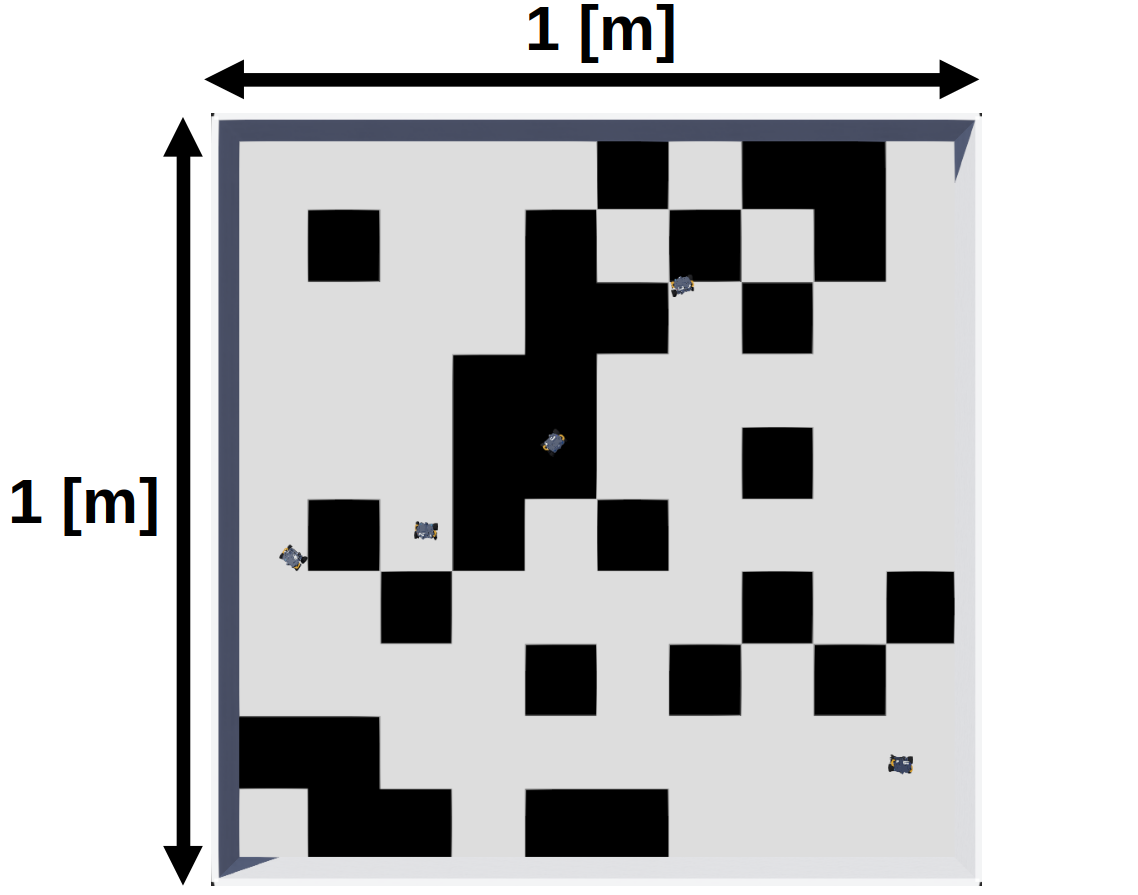}
        \parbox{\linewidth}{\centering (b) Simulation environment}
    \end{minipage}
    \hfill
    \begin{minipage}[t]{0.39\textwidth} 
        \centering
        \includegraphics[width=\linewidth]{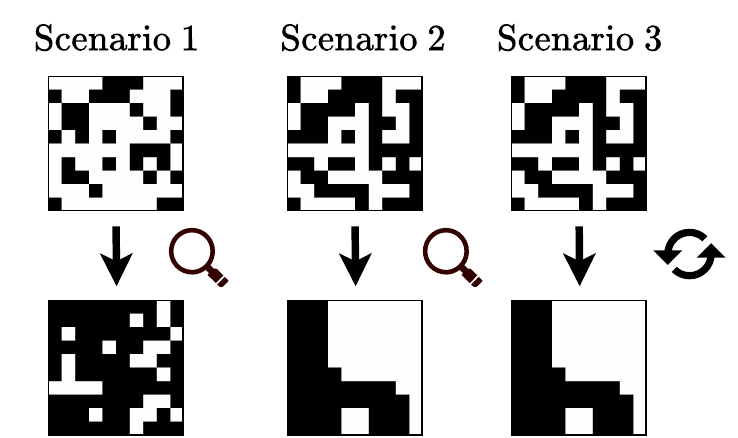}
        \parbox{\linewidth}{\centering (c) Experimental scenarios}
    \end{minipage}
    \caption{Simulations use the vibration-sensing robot (a) within a bounded environment (b). Scenarios (c): fill ratio change detection, spatial-correlation detection, and adaptation to improve accuracy.}
    \label{fig:system_overview}
\end{figure*}

\section{Experimental Framework and Scenarios}
\label{sec:experimental_framework}
We validate our approaches using our previously developed simulation framework, which utilizes a walled environment in the physics-based simulator Webots \cite{Siemensma2026OptimizationRobots}.

\textit{Robot Model}:
Our simulated robot shown in Figure~\ref{fig:system_overview}a is based on \cite{Dementyev2016Rovables:Wearables}, which we previously revised to improve traversability on metal surfaces and update the electronics \cite{Siemensma2024CollectiveInspection}. The model includes sensing and actuation components representative of the physical platform, including differential drive, collision avoidance sensors, and RF24-based communication.
In earlier work, sampling was performed at fixed time intervals \cite{Siemensma2026OptimizationRobots}. Here, we replace this with distance-based sampling using onboard wheel-encoder odometry. Under typical move–sense–move behavior at a speed of 5 cm/s, the battery lifetime of the platform is approximately 1200 s.

\textit{Parameter Selection}: Based on our previously optimized random walk times at 5 cm/s, we rescale them to $R \sim \text{Cauchy}(0.393, 0.536)$ m \cite{Siemensma2026OptimizationRobots}.
Moreover, robots use a uniform prior $\text{Beta}(\alpha_0, \beta_0)$ = $\Betadist(1,1)$, and make final decisions when their belief crosses $p_c = 0.95$. 
We use Python-based Monte Carlo simulations to find the relations for the CUSUM threshold in Equations~\eqref{eq:arl0_selection} and \eqref{eq:h_selection}. We perform 1,000 simulations for each $\theta_s \in {5, 6, \dots, 20}$ at $\delta = 0.05$, and record the corresponding $\text{ARL}_0$ values under a no-change scenario with worst-case variability ($f_0 = 0.5$). This yields the parameters $(c
,h) = (21.10, 0.42)$ in Equations~\eqref{eq:arl0_selection} and \eqref{eq:h_selection}. Moreover, we set $\Delta N = 0.25$ and $\eta = 0.2$, so a robot enlarges its step size when $N_t < 1-\Delta N = 0.75$ and otherwise shrinks it by $\eta\,\Delta N = 5\%$. 

\textit{Baseline Comparison}: Each robot concurrently runs DBB, DBBCPD, and CUSUM to ensure fair comparison by using the same samples. For DBB, the change detection parameters are set to $(p_{\text{lim}}, p_{\text{res}}) = (0.99, 0.55)$, consistent with the original formulation \cite{Pfister2022CollectiveEnvironments}. DBBCPD’s parameters $n_1, n_2, n_3$ depend on swarm density and are therefore rescaled with $N_r$, and capped at $N_r-1$, to preserve equal spatial coverage over time. Additionally, we use the feedback time period of 50 s as in \cite{Pfister2023CollectiveEnvironments}.

\textit{Experimental Scenarios}:
We study three experimental scenarios, averaging results over 100 simulation runs per configuration. 
All experiments run for 1200 s to match the robot battery life, with environmental changes introduced halfway.

\textbf{Scenario 1: fill ratio change detection} We aim to evaluate change-detection performance in scenarios with inverted fill ratios, considering $\Delta f \in \{0.1, 0.2, 0.4\}$. In this scenario, robots use fixed step sizes of $d = \kappa \ell$ at $\ell = 0.1$ m. We study two settings: (i) 5 robots with all-to-all communication, representing a realistic scenario for our robot platform \cite{Siemensma2026OptimizationRobots}, and (ii) 8 robots with communication ranges of three body lengths, providing coverage over time similar to that in the original studies \cite{Pfister2022CollectiveEnvironments, Pfister2023CollectiveEnvironments}. For DBBCPD, setting (i) yields $(n_1,n_2,n_3 )= (2,3,4)$ and setting (ii) yields $(n_1,n_2,n_3 )= (3,4,4)$. For CUSUM, we set $\theta_p = 0.6$ in both settings.

\textbf{Scenario 2: Estimation of correlation} We deploy a swarm of 5 robots with all-to-all communication and aim to estimate sampling correlation in environments with varying tile sizes $\ell \in \{0.05, 0.1, 0.2\}$ m, considering both cases where the step size satisfies $d \leq \ell$ and $d > \ell$. According to our algorithm in Section~\ref{sec:algorithm}, estimation is only feasible when $d \leq \ell$. Additionally, we evaluate how changes in spatial correlation, arising from clustering and tile size changes, affect the rate of false-positive fill ratio change detections under the assumption that the fill ratio remains constant.

\textbf{Scenario 3: Adaptation to correlation} We use the same swarm size and communication settings as in Scenario 2, but now aim to enhance CUSUM's accuracy by studying scenarios where the step size $d$ is either fixed or adaptive. We consider spatial correlations induced by both tile clustering and tile size variations. The threshold $\theta_p$ is set to 0.005, allowing a 0.5\% false-alarm probability assuming ideal sampling.

$\theta_p$ is an experimental operating point, not scenario knowledge the robot needs: Scenario 1 uses a permissive $\theta_p=0.6$ for fast detection, while Scenarios 2--3 hold the fill ratio constant, so a strict $\theta_p=0.005$ suppresses correlation-induced false positives. In deployment, $\theta_p$ is fixed once from the tolerable false-alarm rate and is independent of connectivity, so the robot need not identify the scenario.

\textit{Evaluation Metrics}:
We use four evaluation metrics: accuracy, detection delay, recall, and precision.
First, accuracy is defined as one minus the absolute error between the robot's belief and the true final decision, i.e., $1 - |d_f^* - p|$. Here $d_f^*$ steps at the change point, so $1 - |d_f^* - p|$ is defined at every instant; reported over time, the transition appears as a temporary loss of accuracy.
Second, detection delay is defined as the time between the true change and the first true positive (TP) detection of that change. If no change is detected, the instance is treated as a false negative (FN) and excluded from the delay statistics.
Recall captures the proportion of true changes that are correctly detected, i.e., $\text{Recall} = \frac{\text{TP}}{\text{TP}+\text{FN}}$. Finally, precision captures the proportion of detected changes that correspond to true changes. Low precision therefore indicates high false positive rates, i.e., $\text{Precision} = \frac{\text{TP}}{\text{TP}+\text{FP}}$.
\section{Results and Discussion}
\label{sec:experiments_results}
In this section, we present simulation results for three scenarios: (i) detecting changes in fill ratio, (ii) detecting spatial correlations, and (iii) adapting to spatial correlations.
\begin{figure}[t!]
    \centering
    \begin{minipage}[t]{0.33\textwidth}
        \centering
        \includegraphics[width=\linewidth]{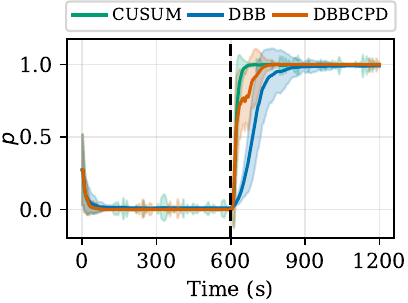}
        (a) $\Delta f= 0.4$, all-to-all comm
    \end{minipage}\hfill
    \begin{minipage}[t]{0.33\textwidth}
        \centering
        \includegraphics[width=\linewidth]{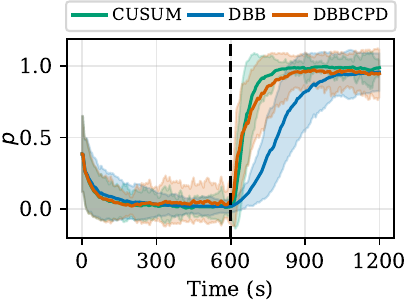}
        (b) $\Delta f= 0.2$, all-to-all comm
    \end{minipage}\hfill
    \begin{minipage}[t]{0.33\textwidth}
        \centering
        \includegraphics[width=\linewidth]{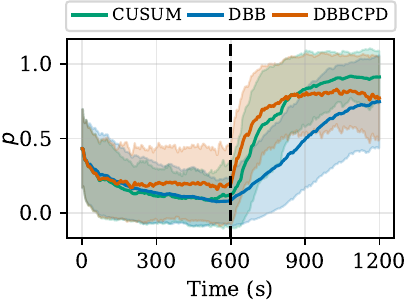}
        (c) $\Delta f= 0.1$, all-to-all comm
    \end{minipage}
    \vspace{.1cm}
    \begin{minipage}[t]{0.33\textwidth}
        \centering
        \includegraphics[width=\linewidth]{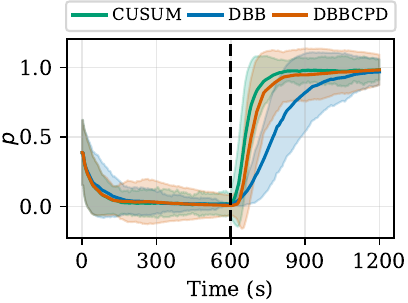}
        (d) $\Delta f= 0.4$, reduced comm
    \end{minipage}\hfill
    \begin{minipage}[t]{0.33\textwidth}
        \centering
        \includegraphics[width=\linewidth]{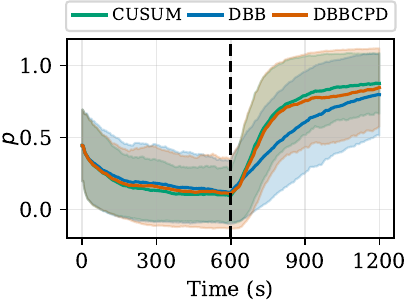}
        (e) $\Delta f= 0.2$, reduced comm
    \end{minipage}\hfill
    \begin{minipage}[t]{0.33\textwidth}
        \centering
        \includegraphics[width=\linewidth]{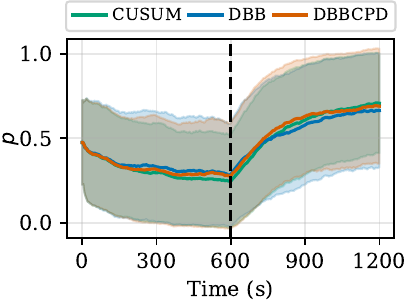}
        (f) $\Delta f= 0.1$, reduced comm
    \end{minipage}
\caption{Belief trajectories (100 runs) for CUSUM ($\theta_p = 0.6$), DBB, and DBBCPD over 1200 s with a fill ratio inversion at $t=600$ s, for all-to-all and reduced (3 body-length range) communication. In easier environments (larger $\Delta f$), CUSUM shows improved belief values over time. In harder environments (smaller $\Delta f$), CUSUM achieves superior beliefs only with increased communication.}
    \label{fig:belief_trajectories}
\end{figure}
\subsection{Scenario 1: Detecting fill ratio Changes}
The belief trajectories of DBB, DBBCPD, and the proposed CUSUM ($\theta_p = 0.6$) are shown in Figure~\ref{fig:belief_trajectories} for changes of $\{(f_0,f_1)\} = \{(0.7,0.3), (0.6,0.4), (0.55,0.45)\}$, evaluated under all-to-all and 3-body-length range communication settings.
First, the belief trajectories of all methods show more accurate pre- and post-change values when communication is increased. For DBB, however, the belief is capped by $p_{\text{lim}}$, a limitation not present in DBBCPD or CUSUM. 
\begin{table}[b!]
\centering
\caption{Evaluation metrics for CUSUM and baseline methods under various fill ratio changes ($\Delta f$) and communication ranges all-to-all and reduced (3-body-length range).}
\label{table:metric_change_detection}
\small
\setlength{\tabcolsep}{2.4pt}
\begin{tabular}{|l|c|c|c|c|c|c|c|c|c|c|}
\hline
\multirow{2}{*}{Range} & \multirow{2}{*}{$\Delta f$}
& \multicolumn{3}{c|}{Precision} 
& \multicolumn{3}{c|}{Recall} 
& \multicolumn{3}{c|}{Delay ($\mu\pm\sigma$) [s]} \\
\cline{3-11}
 & & \scriptsize{CUSUM} & \scriptsize{DBB} & \scriptsize{DBBCPD} 
 & \scriptsize{CUSUM} & \scriptsize{DBB} & \scriptsize{DBBCPD} 
 & \scriptsize{CUSUM} & \scriptsize{DBB} & \scriptsize{DBBCPD} \\
\hline
\multirow{3}{*}{All-to-all}
& $0.4$  & 0.65 & 0.61 & 0.20        & 1.00  & 1.00  & 1.00    & $20\pm6$ & $93\pm21$ & $21\pm9$ \\
& $0.2$  & 0.55 & 0.51 & 0.20        & 0.99 & 0.99 & 0.96      & $49\pm26$ & $197\pm65$ & $50\pm68$ \\
& $0.1$  & 0.52 & 0.33 & 0.13        & 0.88 & 0.80 & 0.98      & $131\pm97$ & $269\pm142$ & $93\pm108$ \\
\hline
\multirow{3}{*}{Reduced}
& $0.4$        & 0.56 & 0.44& 0.97           & 0.98 & 1.00 & 1.00     & $62\pm47$ & $170\pm50$ & $81\pm38$ \\
& $0.2$        & 0.51 & 0.26 & 0.74          & 0.94 & 0.97 & 0.92      & $147\pm122$ & $199\pm116$ & $135\pm99$ \\
& $0.1$        & 0.48 & 0.21 & 0.57           & 0.78 & 0.91 & 0.78     & $214\pm155$ & $185\pm141$ & $199\pm148$ \\
\hline
\end{tabular}
\end{table}
Second, under reduced communication (as considered in baseline studies \cite{Pfister2023CollectiveEnvironments}), CUSUM exhibits similar belief trajectories to DBBCPD (Figure \ref{fig:belief_trajectories}d-f). However, using all-to-all connectivity (Figure \ref{fig:belief_trajectories}a-c), CUSUM's belief is superior due to more frequent false-positive resets in DBBCPD, which lowers its precision ($\sim$2.7--4$\times$ lower than CUSUM, reaching $4\times$ only at $\Delta f=0.1$), as reported in Table \ref{table:metric_change_detection}. Precision depends strongly on connectivity: CUSUM is more precise under all-to-all communication, whereas DBBCPD is more precise under reduced communication (Table 1). CUSUM is nevertheless more consistent across these settings because its nominal false-alarm probability is controlled locally by $\theta_p$. CUSUM's measured pre-change false-alarm rate increased toward the nominal $\theta_p=0.6$ as $f_0$ approached the worst-case calibration point of 0.5, ranging from 0.25–0.54 under all-to-all and 0.43–0.61 under reduced communication.
Neither detector consistently minimizes delay: CUSUM is faster at $\Delta f=0.4$, whereas DBBCPD is faster for smaller changes. Because delay excludes missed detections, it must be interpreted alongside recall; for example, at $\Delta f=0.1$ with all-to-all communication, DBBCPD achieves 0.98 recall versus 0.88 for CUSUM.
Moreover, DBBCPD’s faster detection comes at two costs: (i) precision in Table  \ref{table:metric_change_detection} degrades significantly as $\Delta f$ decreases, whereas for CUSUM only a minor decrease is observed; (ii) DBBCPD stores the full sample stream, requiring $\mathcal{O}(n)$ memory: 82 kB for 600 samples in its original study~\cite{Pfister2023CollectiveEnvironments}, exceeding the 32 kB budget of small robots such as the widely used Kilobot~\cite{Rubenstein2012Kilobot:Behaviors}. Its compute is $\mathcal{O}(n^2)$ from re-running detection over the growing history, whereas CUSUM needs only $\mathcal{O}(1)$ memory ($10 \times 4$ bytes) and per-sample compute via Equations~\eqref{eq:cusum_chart}–\eqref{eq:h_selection}. 
Compared to DBB, CUSUM is substantially faster under all-to-all communication (Table \ref{table:metric_change_detection}, row 3). However, in a reduced-communication setting with $\Delta f=0.1$ (Table \ref{table:metric_change_detection}, row 6), DBB achieves the shortest detection delay. This is likely caused by resets whenever a single robot’s belief drops below $p_{\text{res}}$, consistent with DBB’s very low precision in this scenario.  The CUSUM reference $\hat f_0$ keeps absorbing post-change samples, so it drifts toward the new fill ratio and the statistic stalls; detection power therefore depends on the posterior size ($\alpha+\beta$) at the change.

\subsection{Scenario 2: Detecting Correlations}
\begin{figure}[t!]
    \centering
    \begin{minipage}[t]{0.33\textwidth}
        \centering
        \includegraphics[width=\linewidth]{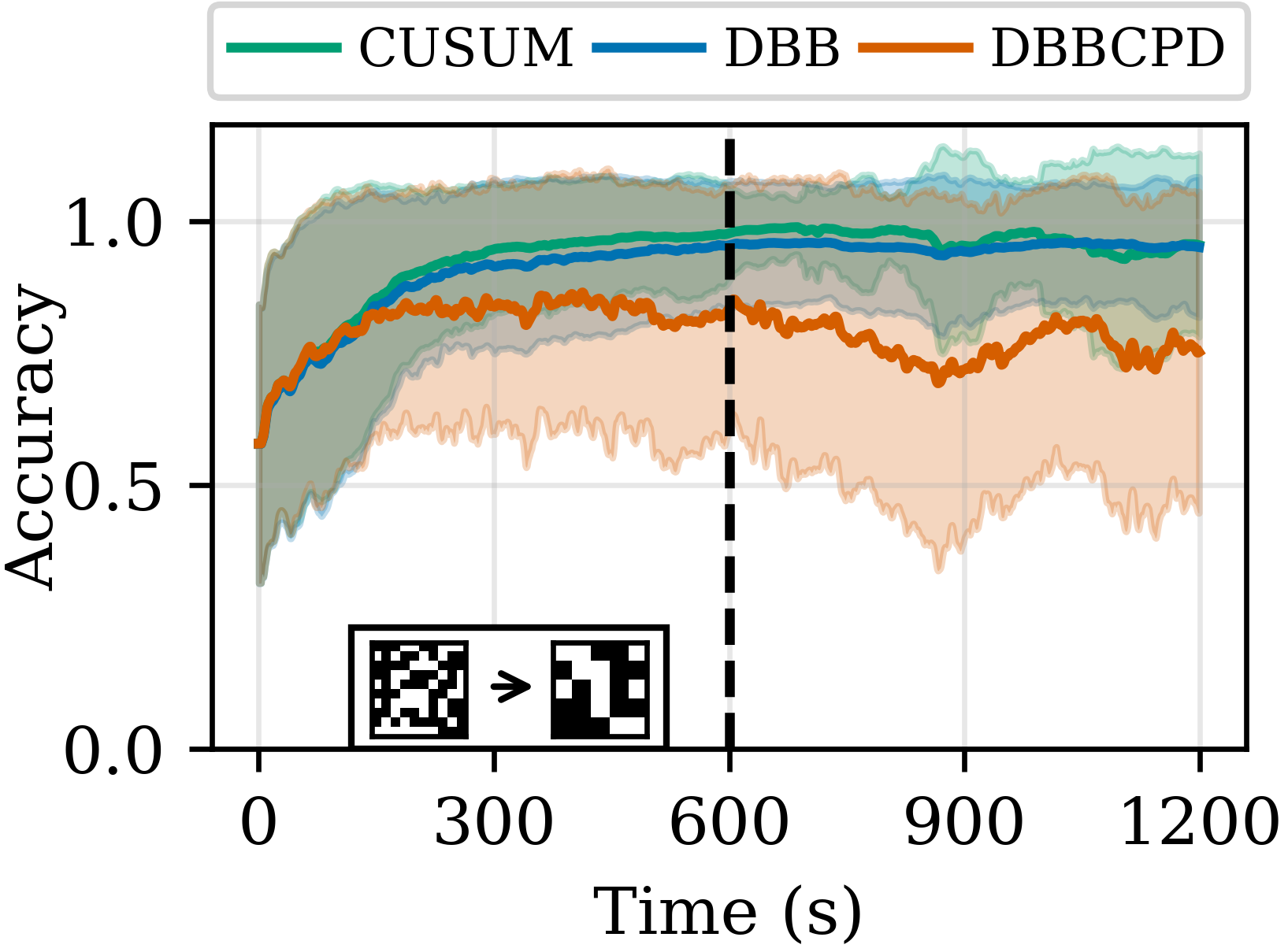}
        (a) Low correlation
    \end{minipage}\hfill
    \begin{minipage}[t]{0.33\textwidth}
        \centering
        \includegraphics[width=\linewidth]{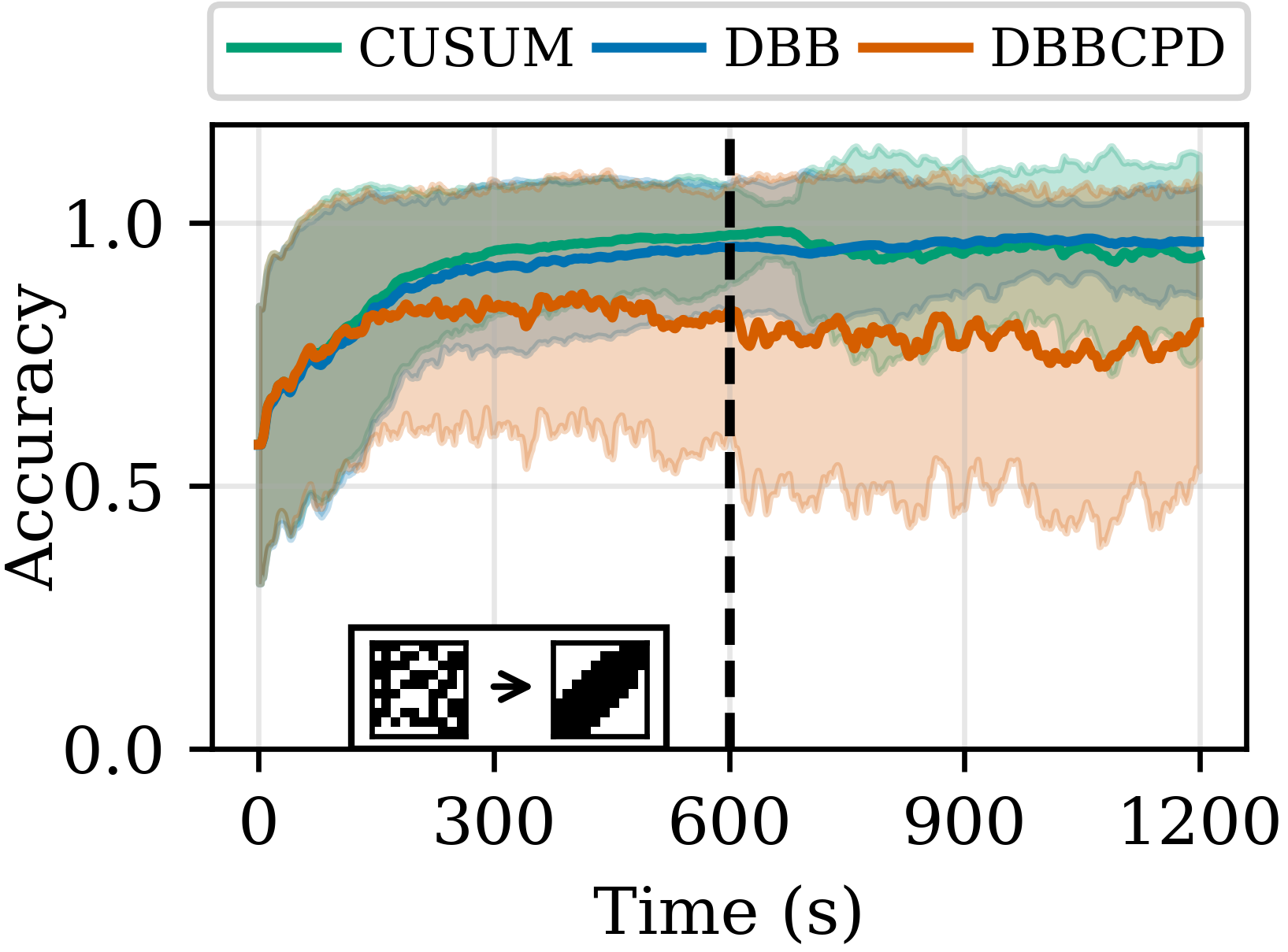}
        (b) Medium correlation
    \end{minipage}\hfill
        \begin{minipage}[t]{0.33\textwidth}
        \centering
        \includegraphics[width=\linewidth]{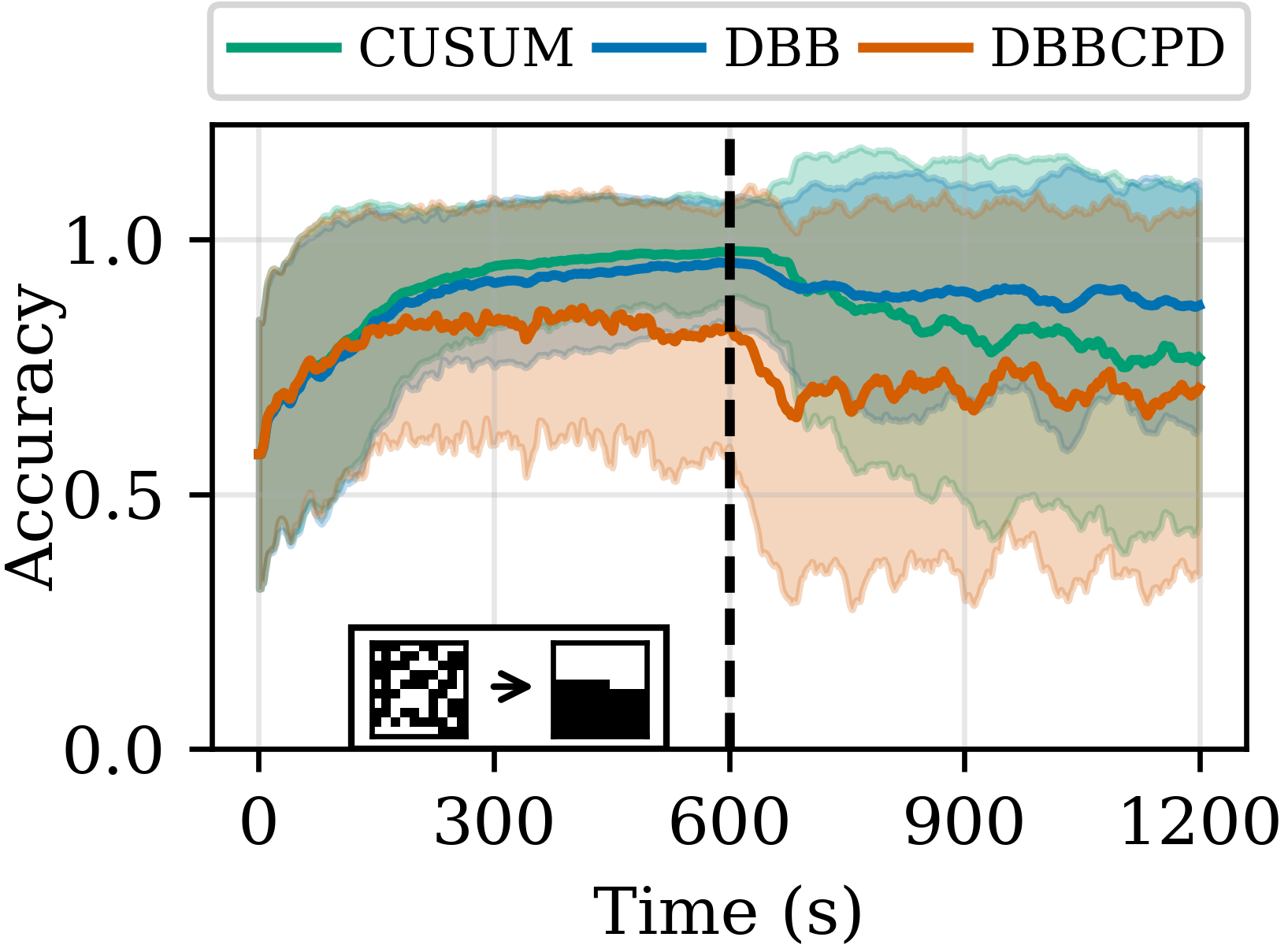}
        (c) High correlation
    \end{minipage}\hfill
    \caption{Accuracy ($1 - |d_f^* - p|$) at fixed step sizes proportional to the pre-change correlation, for CUSUM ($\theta_p = 0.005$) and the baseline methods DBB and DBBCPD, in a scenario where the fill ratio ($f = 0.56$) remains constant but spatial correlation changes at $t = 600$ s. All methods exhibit accuracy degradation as correlation increases, likely caused by false-positive fill ratio detections.}
    \label{fig:corr_effect}
\end{figure}

Our second result focuses on the effect and detection of spatial correlations. First, we compare CUSUM ($\theta_p=0.005$) against DBB and DBBCPD in scenarios where the spatial correlation changes ``random ($\ell=0.1$ m)$\rightarrow$random ($\ell=0.2$ m)'', ``random$\rightarrow$diagonal'', and ``random$\rightarrow$stripe'' at fixed $f=0.56$. 
The results in Figure~\ref{fig:corr_effect} indicate an accuracy loss induced by the change for all methods. Notably, DBB exhibits the smallest accuracy loss as spatial correlation increases. Its inferior detection delay in Scenario 1 appears to improve accuracy in this scenario through the detection delay–accuracy trade-off. In contrast, CUSUM's accuracy degrades after the change and does not recover over time, although its performance is expected to improve by lowering $\theta_p$, i.e., fewer false-positive detections. DBBCPD performs even worse as all-to-all communication causes excessive false-positive resets, thereby lowering accuracy even before the correlation change. Overall, changes in correlation highlight the speed-accuracy trade-off, where faster fill ratio detections cause lower accuracy in correlated environments, which is tunable for CUSUM.

\begin{figure}[t!]
    \centering
    \begin{minipage}[t]{0.33\textwidth}
        \centering
        \includegraphics[width=\linewidth]{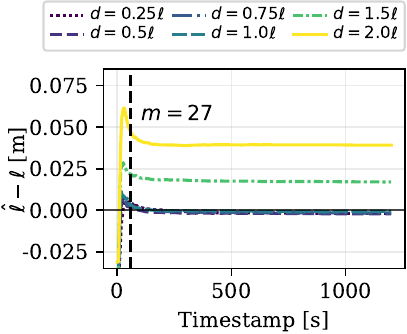}
        (a) $\ell = 0.05$ m
    \end{minipage}\hfill
    \begin{minipage}[t]{0.33\textwidth}
        \centering
        \includegraphics[width=\linewidth]{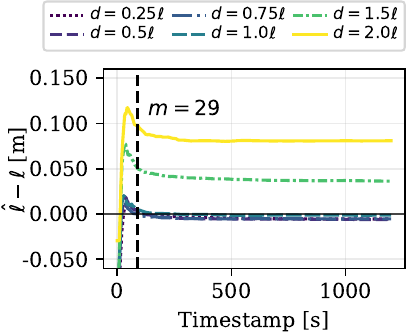}
        (b) $\ell = 0.1$ m
    \end{minipage}\hfill
        \begin{minipage}[t]{0.33\textwidth}
        \centering
        \includegraphics[width=\linewidth]{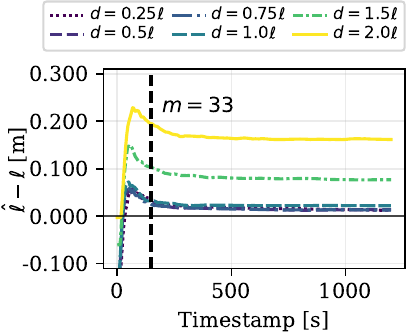}
        (c) $\ell = 0.2$ m
    \end{minipage}\hfill
\caption{Average tile-size estimates over 100 runs with five robots for $\ell \in {0.05, 0.1, 0.2}$ m. Estimates converge near $\ell$ for $d \leq \ell$ and toward $d$ otherwise. Dashed lines indicate the mean convergence point m within 10\% of steady-state.}
    \label{fig:ell_estimates}
\end{figure}

In addition, we aim to improve accuracy by letting robots adapt their step size $d$ every $m$ samples. To determine a suitable value for $m$, we evaluate the tile-size estimation error, $\hat{\ell} - \ell$, over time for different combinations of $d$ and $\ell$. Figure~\ref{fig:ell_estimates} shows that for $d \leq \ell$ the estimator converges to the true tile size, whereas for $d > \ell$ a steady-state bias appears that is proportional to $d$, as expected per Equation \eqref{eq:hat_ell}.
Moreover, for $\ell = 0.2$ (Figure~\ref{fig:ell_estimates}c), the estimator exhibits a slight positive bias due to increased spatial correlations induced by boundary reflections of the random walk, leading to repeated sampling of boundary tiles, consistent with the corner-dense sample distribution reported for this platform~\cite{Siemensma2026OptimizationRobots}. Finally, we compute the average value of $m$ at which the estimator reaches within 10\% of its steady state, and find it increases slightly with tile size. Since our experiments in Scenario 3 are conducted using a pre-change tile size $\ell = 0.1$, we select $m = 29$.


\subsection{Scenario 3: Adapting to Correlations}
\begin{figure}[t!]
    \centering
    \begin{minipage}[t]{0.33\textwidth}
        \centering
        \includegraphics[width=\linewidth]{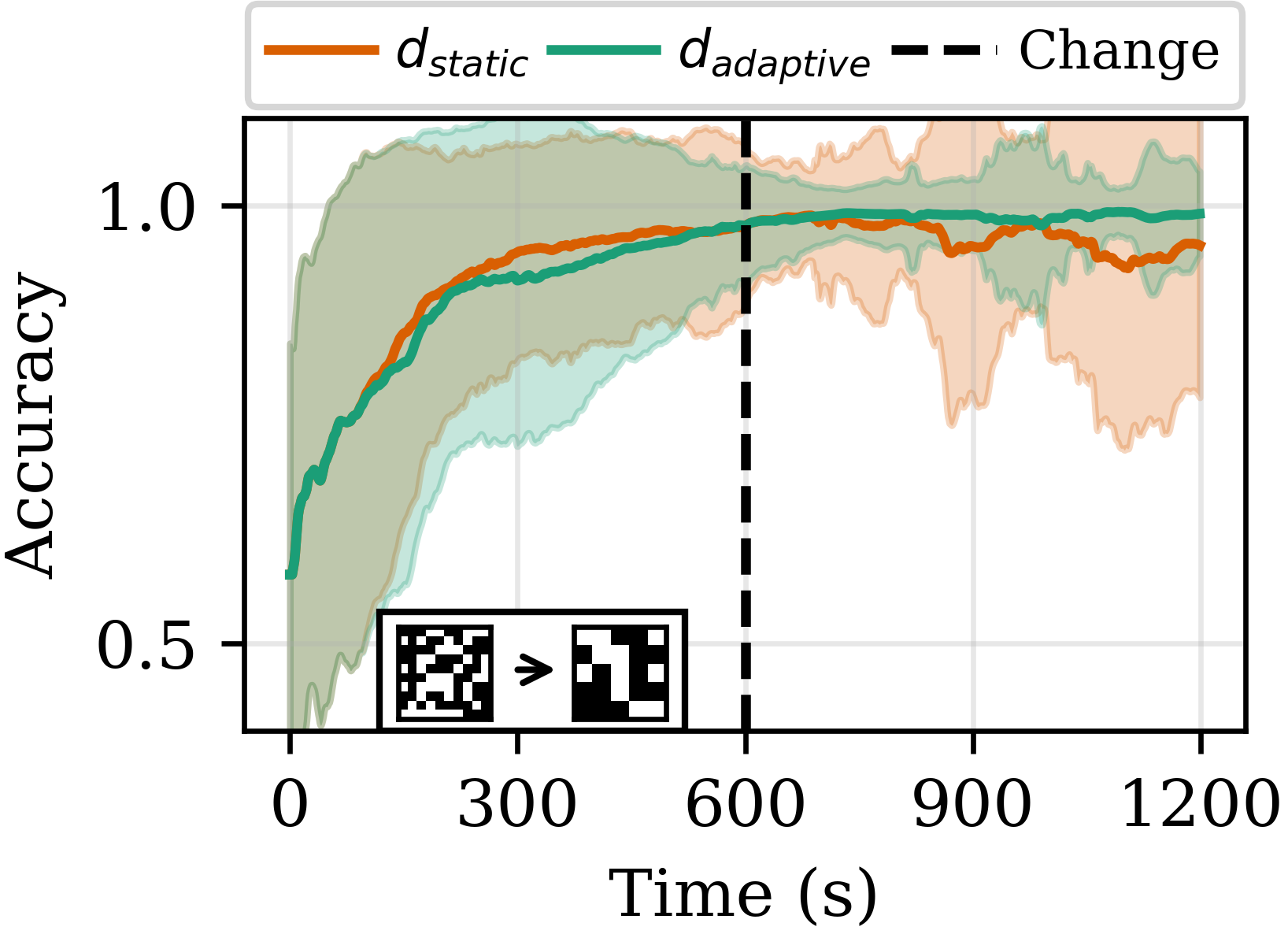}
        (a) Low correlation
    \end{minipage}\hfill
    \begin{minipage}[t]{0.33\textwidth}
        \centering
        \includegraphics[width=\linewidth]{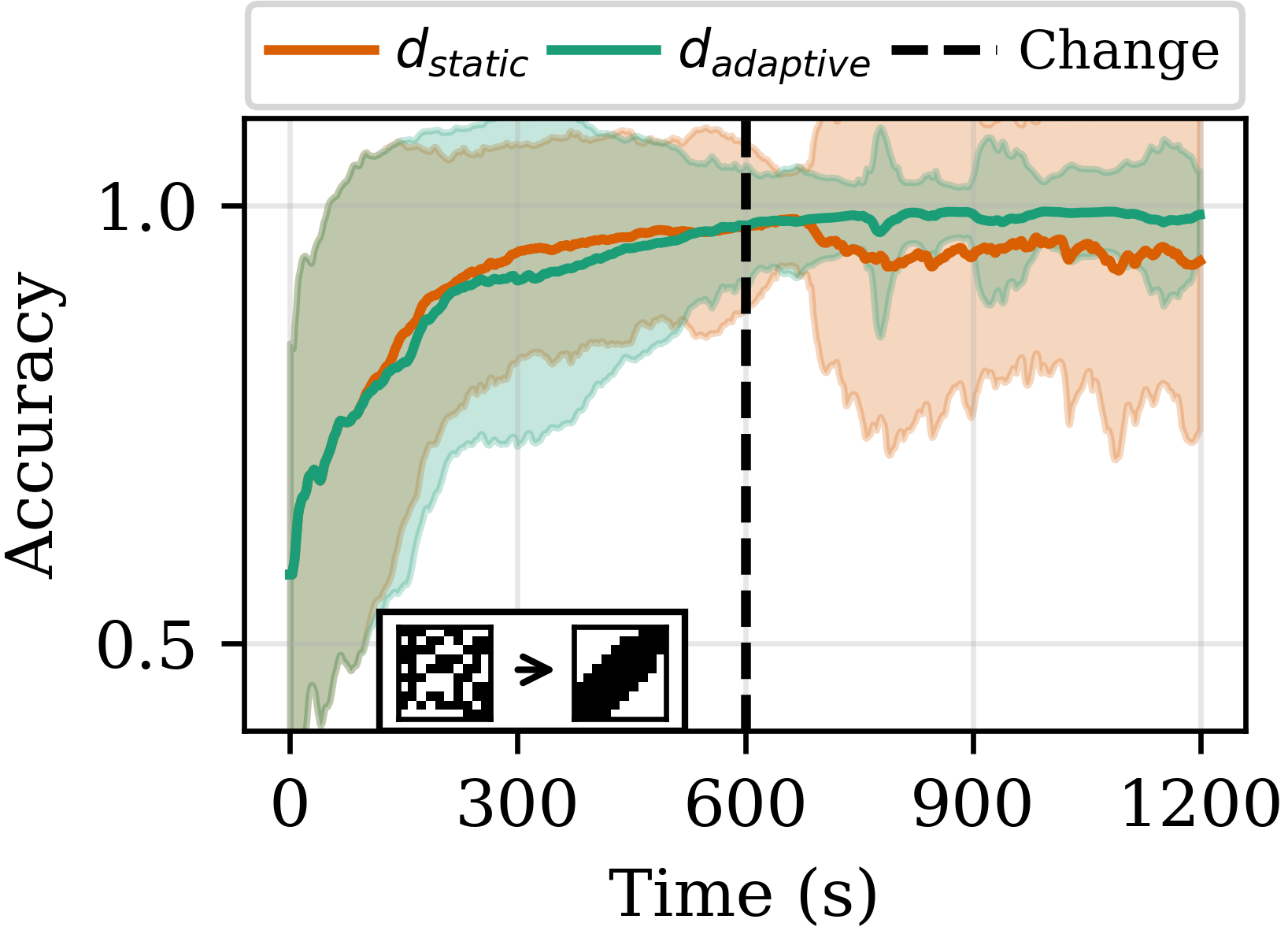}
        (b) Medium correlation
    \end{minipage}\hfill
    \begin{minipage}[t]{0.33\textwidth}
        \centering
        \includegraphics[width=\linewidth]{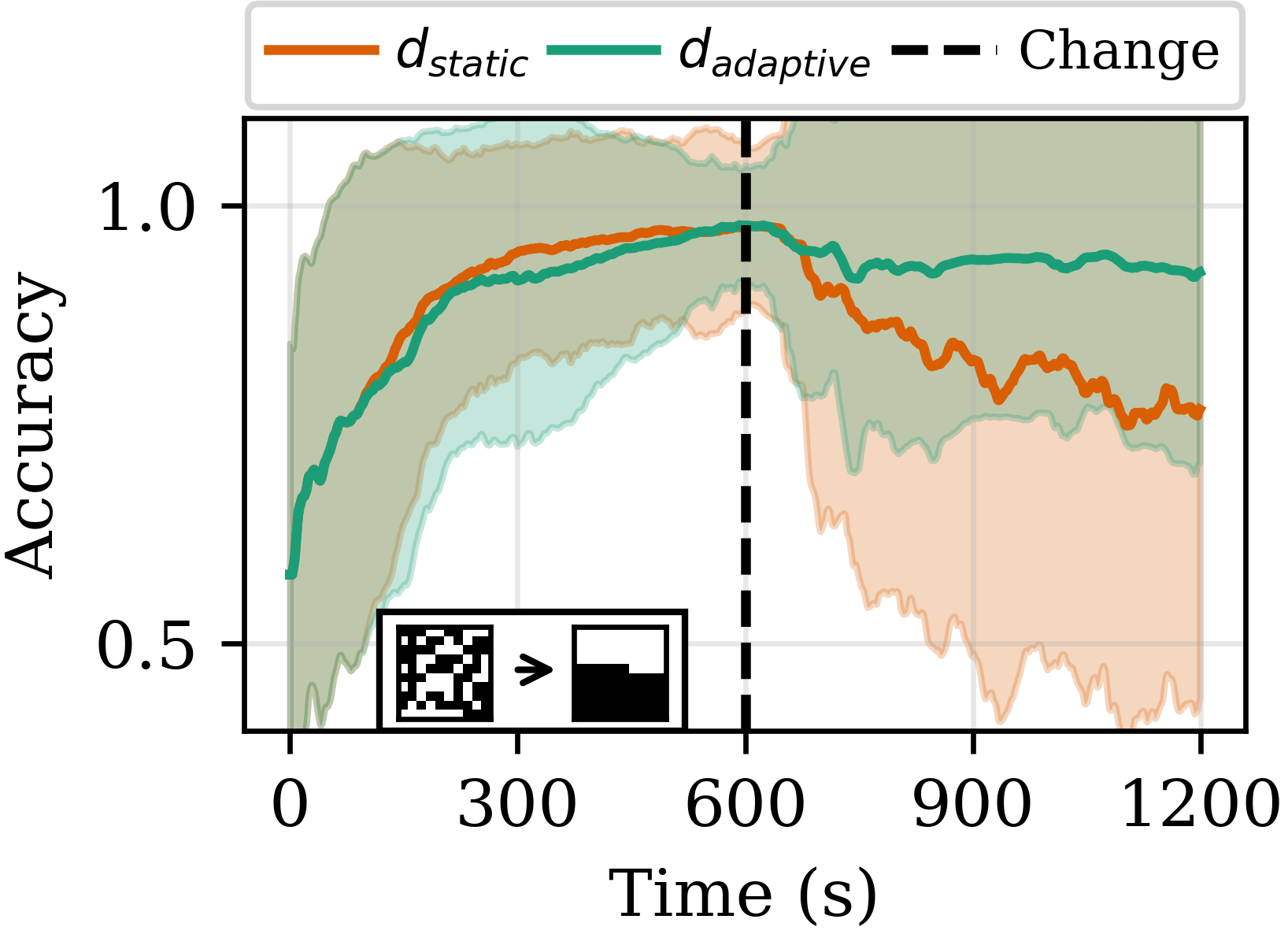}
        (c) High correlation
    \end{minipage}
    \caption{Accuracy of the proposed CUSUM ($\theta_p = 0.005$) for fixed and adaptive step size $d$ across three environments with varying degrees of spatial correlation, while keeping the fill ratio constant ($f = 0.56$).
    Overall, CUSUM exhibits an increasing number of false positives as spatial correlation strengthens, due to correlated sampling.
    Adapting the step size based on the observed correlation enables more representative sampling, thereby reducing false positives.}
    \label{fig:correlation_plots}
\end{figure}

Here we examine adaptation to spatial correlations via the step size $d$ in order to maintain high decision accuracy. We consider the same environmental transitions (random (small) $\rightarrow$ random (big)), (random $\rightarrow$ diagonal), and (random $\rightarrow$ stripe), and compare fixed step sizes against an adaptive strategy (initialized at $d= \kappa\ell$) in the proposed CUSUM ($\theta_p = 0.005$). The results in Figure~\ref{fig:correlation_plots} show that adapting $d$ increases decision accuracy by reducing falsely detected fill ratio changes. This demonstrates that spatial correlations can be effectively mitigated through correlation estimation and step-size adaptation. Notably, accuracy improves even when the CUSUM threshold $\theta_s$ is lower in the adaptive case, as the increased step sizes reduce the sample rate and thereby the threshold $\theta_s$ through Equations \eqref{eq:arl0_selection} and \eqref{eq:h_selection}. Overall, actively estimating and adapting to spatial correlation enhances decision-making accuracy, particularly under strong correlations (Figure~\ref{fig:correlation_plots}c).

\section{Conclusion and Future Work}
\label{sec:discussion_conclusion}

We addressed collective perception in dynamic environments with changing fill ratios and spatial correlations. We introduced (i) an $\mathcal{O}(1)$-memory CUSUM fill ratio change detector, (ii) an online estimator of sampling correlation, and (iii) adaptive distance-based sampling to mitigate sampling correlation. CUSUM achieves competitive detection delays and more consistent precision across communication settings while remaining suitable for resource-constrained robots. Adaptive sampling substantially improves decision accuracy under strong spatial correlations.

The correlation estimator cannot identify excessive de-correlation and may confound environmental correlation with persistent-walk and boundary effects, while CUSUM can degrade as its reference absorbs post-change samples. Future work will investigate improved CUSUM references, simultaneous fill ratio and correlation changes, and real-world validation including wheel-odometry slip and drift.

\bibliographystyle{spmpsci}   
\bibliography{references-v1}
\end{document}